# CNN based Intelligent Streetlight Management Using Smart CCTV Camera and Semantic Segmentation

Md Sakib Ullah Sourav[1], Huidong Wang[1], Mohammad Raziuddin Chowdhury[2], Rejwan Bin Sulaiman[3]

[1]School of Management Science and Engineering, Shandong University of Finance and Economics, Jinan, Shandong, China
[2]Department of Statistics, Jahangirnagar University, Bangladesh
[3]Northumbria University, United Kingdom

Emails: sakibsourav@outlook.com, huidong.wang@ia.ac.cn, razichy3@gmail.com, rejwan.binsulaiman@gmail.com

**Abstract**

One of the most neglected sources of energy loss is streetlights that generate too much light in areas where it is not required. Energy waste has enormous economic and environmental effects. In addition, due to the conventional manual nature of operation, streetlights are frequently seen being turned 'ON' during the day and 'OFF' in the evening, which is regrettable even in the twenty-first century. These issues require automated streetlight control in order to be resolved. This study aims to develop a novel streetlight controlling method by combining a smart transport monitoring system powered by computer vision technology with a closed circuit television (CCTV) camera that allows the light-emitting diode (LED) streetlight to automatically light up with the appropriate brightness by detecting the presence of pedestrians or vehicles and dimming the streetlight in their absence using semantic image segmentation from the CCTV video streaming. Consequently, our model distinguishes daylight and nighttime, which made it feasible to automate the process of turning the streetlight 'ON' and 'OFF' to save energy consumption costs. According to the aforementioned approach, geo-location sensor data could be utilized to make more informed streetlight management decisions. To complete the tasks, we consider training the U-net model with ResNet-34 as its backbone. Validity of the models is guaranteed with the use of assessment matrices. The suggested concept is straightforward, economical, energy-efficient, long-lasting, and more resilient than conventional alternatives.

## 1 Introduction

The street lighting system is a crucial component of modern society, providing a range of important benefits. Students have a greater sense of safety, evening vehicular movement is improved, nocturnal accident risks are reduced, properties may be better safeguarded, and there is less chance of damage to property [1]. Even though the benefits of streetlights are obvious, it is important to keep in mind that this technology places a huge load on the nation's energy budget due to its large energy consumption burden. Efficient street lighting systems are needed to tackle energy crisis nowadays because they can help reduce energy consumption and conserve energy sources. Streetlights can consume vast amounts of energy, which adds to the cost of providing light to street-level areas. This potentially leads to a significant impact on the amount of energy consumed, particularly in urban areas. By using efficient street lighting system, the amount of energy required to provide lighting can be greatly reduced. This, in turn, can help reduce the amount of energy consumed and save on energy costs. In addition, efficient streetlights can also help reduce the amount of carbon emissions, thus helping to fight climate change. Such efficient street lighting systems also help to provide a better quality of light, which can improve visibility in the dark and help to reduce the risk of accidents. Additionally, cities can also save money on maintenance costs for their streetlights, as they will be able to use less energy to power them, and replace their bulbs less often. By providing better lighting, cities get to improve their public safety, as this can help to reduce the amount of crime that takes place at night.

In recent years, as the need for a solution to the worldwide energy problem has been more urgent, the researchers have made continual attempts to construct a robust street lighting system that may limit the amount of electricity that is consumed. One of the additional goals of these efforts is to reduce the amount of carbon dioxide emissions that are caused by the production of electricity. A prototype was built by Abdullah and his colleagues [2] that makes use of a Light Dependent Resistor (LDR), an Infrared sensor (IR), a battery, and an LED in order to automatically alter the brightness of the LED so that electricity can be saved. Mary et al. [3] developed a solution that was quite similar, but it had the additional advantage of being an intelligent street lighting system. This system made use of sensors and a controller to operate. In their study, Islam et al. [4] recommended the use of infrared (IR) sensors to detect vehicle movement and pedestrians, in addition to having the capacity to identify road accidents and tell the proper authorities with the location and number of the automobile involved in the collision. Arun et al. [5] devised a method that saves energy by shutting off street lighting during the daytime by employing PIR and LDR sensors to activate a cluster of lights surrounding vehicles. Other attempts [6-11] those were analogous to this one made use of a wide array of sensors, detectors, and microcontrollers, all of which were connected to the Internet of Things. (IoT).

Despite these developments, there is still a lack of smart streetlights that can develop an energy-efficient system with fewer physical design complexities and low maintenance labor and expense. To address this need, this current study looks to semantic segmentation, more commonly referred to as image segmentation, to help create an efficient streetlight management system. Semantic segmentation is the process of grouping portions of an image that belong to the same item class using various convolutional neural networks (CNNs). Semantic segmentation is a game changer in computer vision-based problems because it allows for more accurate object detection and recognition. Unlike traditional computer vision algorithms, which assign a single label to each object in an image, semantic segmentation assigns a label to each pixel in the image. This makes it easier to detect and recognize complex objects, such as cars, buildings, and trees, by distinguishing between different object parts. Additionally, semantic segmentation enables models to identify and segment objects even when they are partially occluded, which helps computers better understand the environment around them. This is especially useful for self-driving cars, where accurate object detection and recognition is essential for safety. This approach has been successfully employed in a variety of applications, such as localization [12-14], scene interpretation [15-17], robotic navigation [18], fire detection in surveillance systems [19-22], roadside occupancy surveillance systems [23], and autonomous vehicle driving [24-26]. To the best of our knowledge, semantic segmentation has not yet been applied holistically in tackling the issues related to efficient streetlight management. In many cases, streetlight operation is still performed manually, leading to inefficient and biased practices. Streetlights are often found to be "ON" before evening and "OFF" late at night, despite the clear energy waste this causes. This study looks to evaluate the effectiveness of semantic segmentation in resolving this issue.

The proposed technique, as discussed in Section 2, revolves around the use of semantic segmentation to detect motion and identify the presence of vehicles and pedestrians in the area. This will enable the street lighting system to adjust its settings automatically, with lights turning on and off as needed. This will result in reduced energy consumption and improved streetlight management. The results of the study suggest that semantic segmentation can be an effective tool for efficient streetlight management. The data showed that the proposed technique was able to accurately detect vehicles and pedestrians in the area, allowing the street lighting system to adjust its settings accordingly. This resulted in a significant decrease in the amount of energy consumed by the streetlights, and an overall improved performance of the system. This study has demonstrated the effectiveness of semantic segmentation in streetlight management. The proposed technique is expected to able to provide an accurate and reliable method for detecting motion, which allowed for efficient energy consumption and improved streetlight management. It is hoped that this research will provide a valuable framework for the development of future streetlight systems, leading to improved energy efficiency and better streetlight management. The results indicated that this technique could be an effective tool for efficient streetlight management, resulting in reduced energy consumption and improved system performance. It is hoped that this research will provide a valuable framework for future streetlight systems, leading to improved energy efficiency and better streetlight management.

The subsequent sections of the paper are structured as follows. The section 2 of this article elaborates on the proposed methodology. Section 3 discusses results. Section 4 has the challenges and discussions. Limitations of implementing such frameworks have been discussed in Section 5 and in Section 6, a concluding statement is provided.

## 2 The Proposed Framework

Recent advancements in embedded processing capabilities and the potential of deep features in transportation surveillance activities, such as traffic lane detection [31, 32], vehicle number plate detection [33, 34], traffic flow monitoring [35], roadside occupation detection [36], etc. have motivated us to analyze CNNs and to incorporate the automation of streetlight control. This is done in order to minimize the power usage by employing a semantic segmentation technique within the same framework.

Our architecture for integrating automatic streetlight control with a closed-circuit video (CCTV) network-based intelligent transport management system is illustrated in Fig. 1 (a).

The core idea of our proposed architecture is to capture and process the video frames from the CCTV network in order to detect the presence of vehicles. This is done through a CNN-based object detection technique. Once the object is detected, the semantic segmentation technique is used to identify the exact area of the vehicle. Based on this segmentation, the streetlight can be automatically switched on or off depending on the presence of the vehicle.

In order to facilitate the seamless integration of the automatic streetlight control with the CCTV network-based system, we developed a multi-agent system (MAS) architecture. The MAS architecture consists of various agents that can be used to detect the presence of vehicles, identify the exact location of the vehicle, and control the streetlight. The agents in the MAS architecture are designed to be distributed, cooperative, and dynamic. This helps to reduce the workload of the CCTV network and to ensure that the streetlight control is efficient and reliable.

The proposed architecture for the automatic streetlight control is expected to provide a better way to manage the power usage and to reduce the workload of the CCTV network. By combining the object detection, semantic segmentation, and the MAS architecture, our proposed architecture can provide a more efficient and reliable way to control the streetlight. This method can help to reduce the power usage and to improve the safety and security of the transportation system.

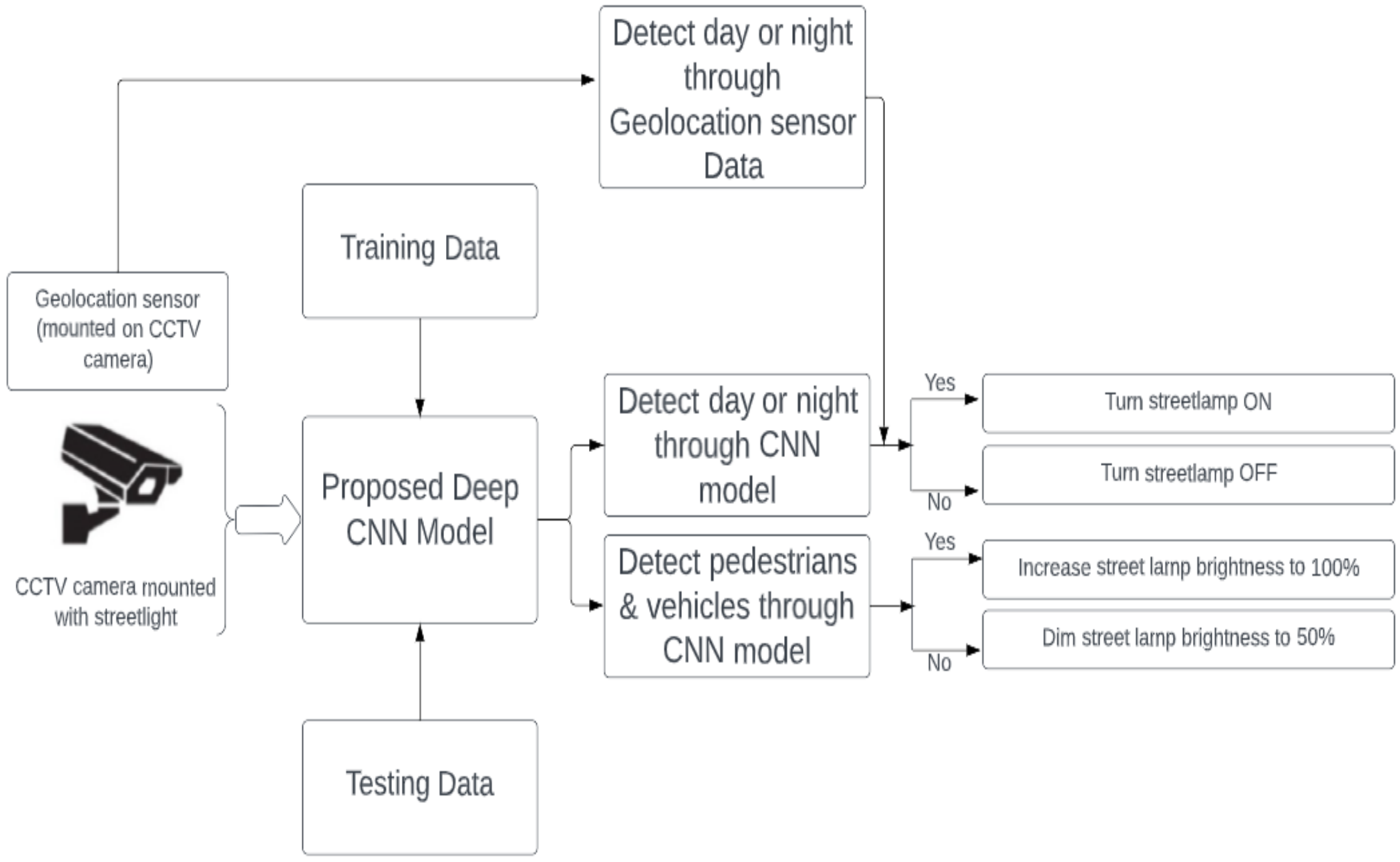

(a)

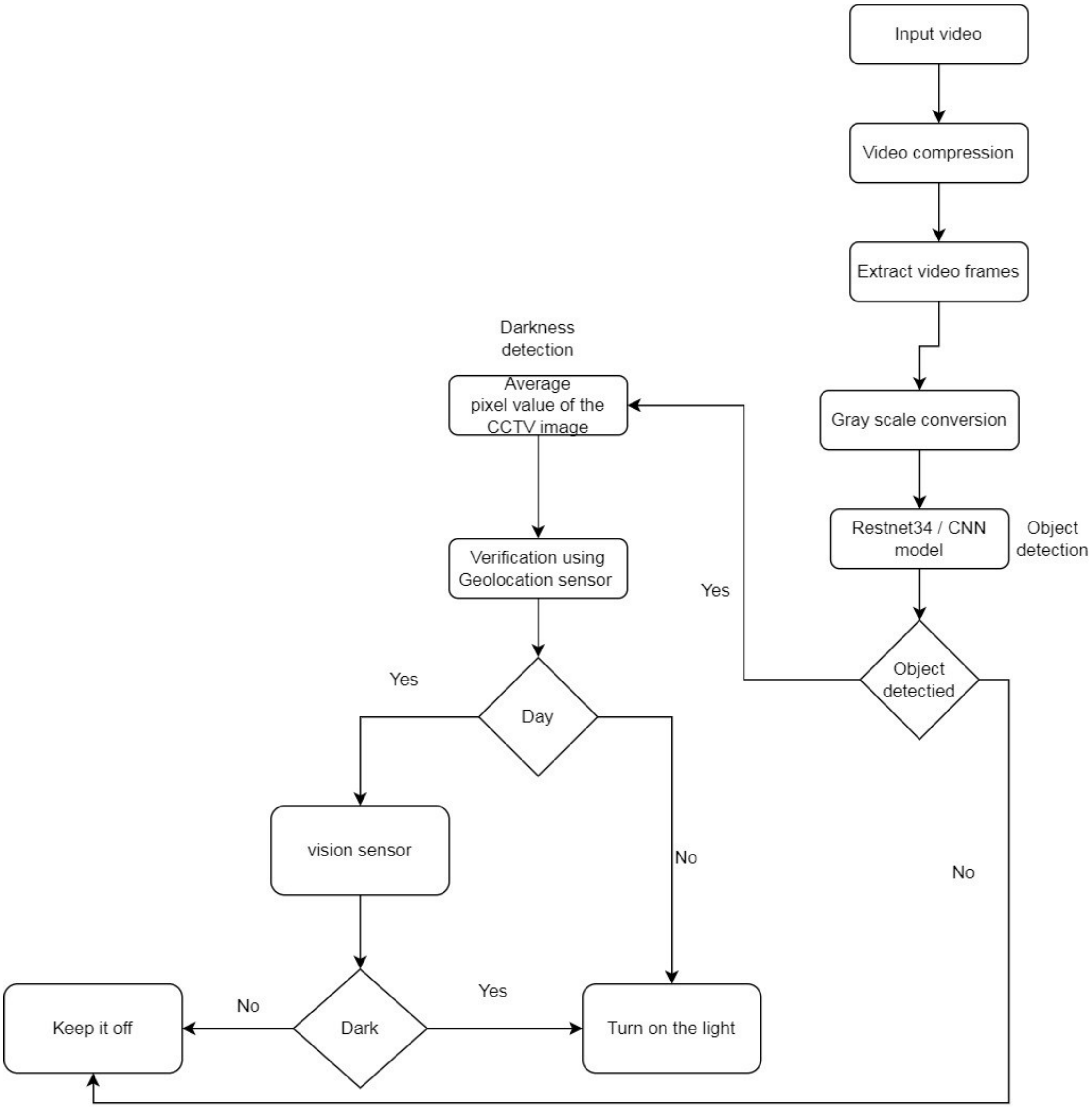

(b)

**Fig.1** Proposed framework overview (a) and step-by-step working principle (b) of automatic streetlight controlling using semantic segmentation.

In our proposed architecture, the trained CNN model is used to recognize pedestrians and cars from the live video stream of CCTV camera. Whenever the model detects the target items, it adjusts the streetlight's brightness to 100 percent. In the opposite situation, the brightness is reduced to 50 percent or less. The algorithm also estimates the brightness of the day based on the pixel values of images collected from the live video stream.

In addition to the model's prediction, we recommend having a geo-location sensor on the CCTV of street lamps. The predictions of the algorithm may not be accurate due to varying illumination conditions during wet weather and abnormalities in the CCTV cameras' lenses. Therefore, location-aware geo-location sensors can further enhance the implementation. If the area under discussion is taken into account carefully, the sensor could be installed to cover a wide area. Inexpensive geological sensors like GPS and wireless weather sensors can also reduce the cost of such installations. To further ensure if it is day or not, vision sensors may also be employed, as demonstrated in Fig. 1(b).

The internal logic of the system will turn ON the streetlights when the brightness is below a certain level. On the other hand, if the brightness is above the threshold, the lamps will be turned OFF. The algorithm is able to adjust the brightness accordingly and assure the safety of pedestrians and drivers. In addition, the geo-location sensors can provide more accurate and reliable information about the day and night time. With this system in place, the streets will be well illuminated, and the safety of pedestrians and drivers will be improved.

## 2.1 Dataset

We propose the use of two datasets to train two separate models for our framework. For the detection task of pedestrians and vehicles, Cambridge-driving Labelled Video Database (CamVid)[1], which is the first collection of videos with object class semantic labels, complete with metadata (Fig.2).

The database provides ground truth labels that associate each pixel with one of 32 semantic classes including animal, bicyclist, car, road, pedestrian, sidewalk, etc. There are over 700 images that were specified manually for the per-pixel semantic segmentation task. For the day and night detection task, we propose the use of a Kaggle dataset[1] named "daytime and night time road images" which comprises of 14,607 day light images and 16,960 nighttime images.

The pseudo code for proposed models is included in Algorithm 1. A trained CNN model is used to recognize people and vehicles from the live CCTV footage video and compress it. Some pre-processing steps are carried out afterwards like, frame extractions, gray scale conversion and others. When the model detects any of the objects of its interest, the streetlight brightness is adjusted to 100%. When the model does not detect any of these, the brightness is reduced to 50% or below. The algorithm also assesses the ambient brightness from the images of the video stream. In addition to the model's predictions, we suggest that the CCTV of the street lamps should have a geo-location sensor. The algorithm forecasts may not be accurate due to varying light conditions during wet weather and any irregularities in the CCTV cameras. This sensor can cover a vast area and make the implementation more efficient. Additionally, low-cost geological sensors like GPS and wireless weather sensors can help reduce the cost of installation. To further confirm if it is day or night, vision sensors can also be used. The internal logic of the system will turn ON the streetlights when the brightness is below a specific level. If the brightness is above the threshold, the lamps will be put OFF. The algorithm can accordingly adjust the brightness and ensure the safety of drivers and pedestrians. The geo-location sensors can provide more precise and dependable information regarding day and night time. With this system in place, the streets will be well illuminated and the safety of drivers and pedestrians will be enhanced. Finally the model accomplishes the task by converting the label into given categories.

[1] https://www.kaggle.com/datasets/raman77768/day-time-and-night-time-road-images
[1] https://www.kaggle.com/datasets/carlolepelaars/camvid

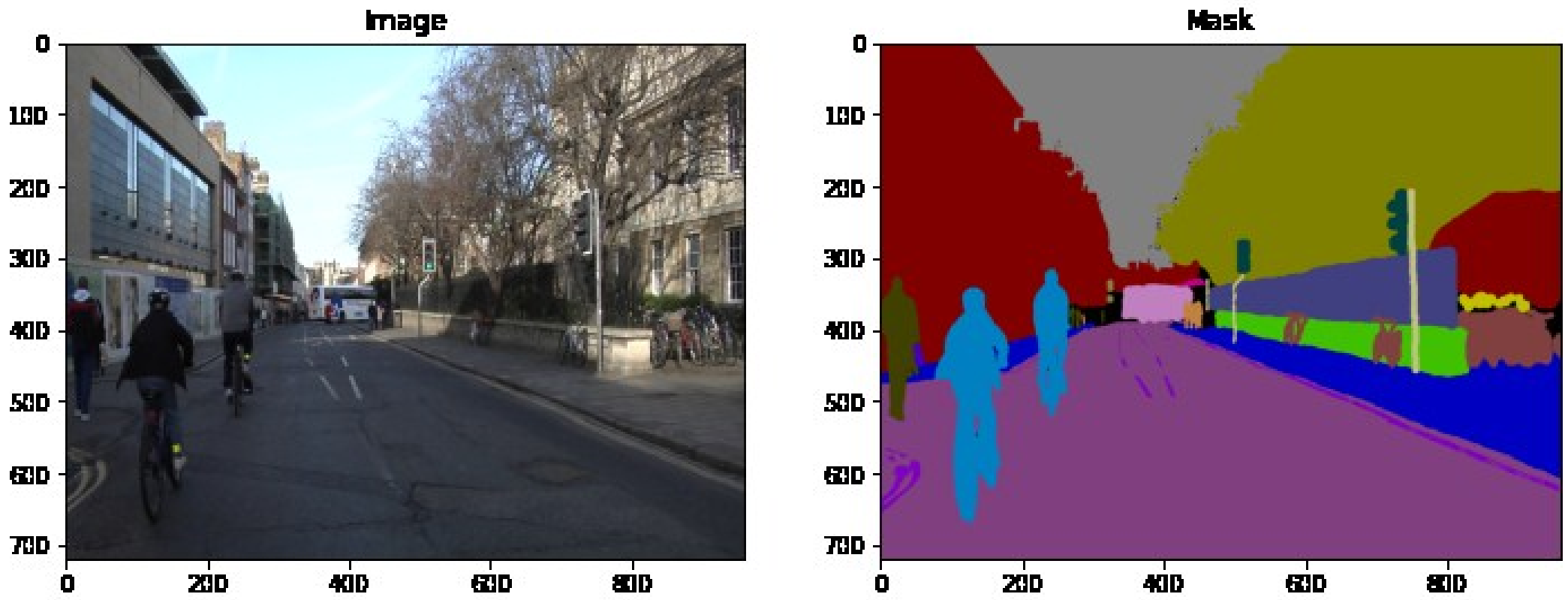

**Fig.2** Sample image and its corresponding masked image from CamVid dataset

**Algorithm 1. Pseudo-Code of proposed model**

```
Input: Video data
Output: Pre-processed compressed video
    for x ← 1 to n do    # x = video
        Label= Split label from the sample and store as matrix
        for image ← 1 to x do
            Call Pre-process image function
            Ri ← input video
            Ci ← Video compression
            Cj← Extract video frames
            Ei ← Gray scale conversion
            Ri← Object (pedestrian/vehicle) detection
            Rj ←Removing noise
            Rs ← Darkness (day/night) detection
            D ← Append image data
        End
        Cd ← convert data into matrix
        Cl ← Convert label into categories
    End
End of Pseudo-Code
```

### 2.2 Detecting Pedestrians and Vehicles

U-net architecture is a type of end-to-end network based on a fully convolutional network that is used for the detection of pedestrians and automobiles (Fig.3). This architecture consists of an encoder and decoder section with skip links. This architecture is especially useful for exact segmentation and pinpointing the location of high-resolution features [27]. The skip links combine the input of each encoding step with the input of its corresponding decoding stage. The encoder component of this architecture is based on ResNet-34 and collects features using multiple convolutions and ReLU activations. This helps in compressing the features and reducing the size of the network. The decoder component of U-net architecture is responsible for decompressing these compressed features using de-convolutions and ReLU activations. This helps in restoring the high-resolution features. Overall, U-net architecture is a powerful end to end network that is used for the detection of pedestrians and automobiles. It helps in exact segmentation and pinpointing the location of high-resolution features. This architecture consists of an encoder component based on ResNet-34 and a decoder component that decompress the features through de-convolutions and ReLU activations.

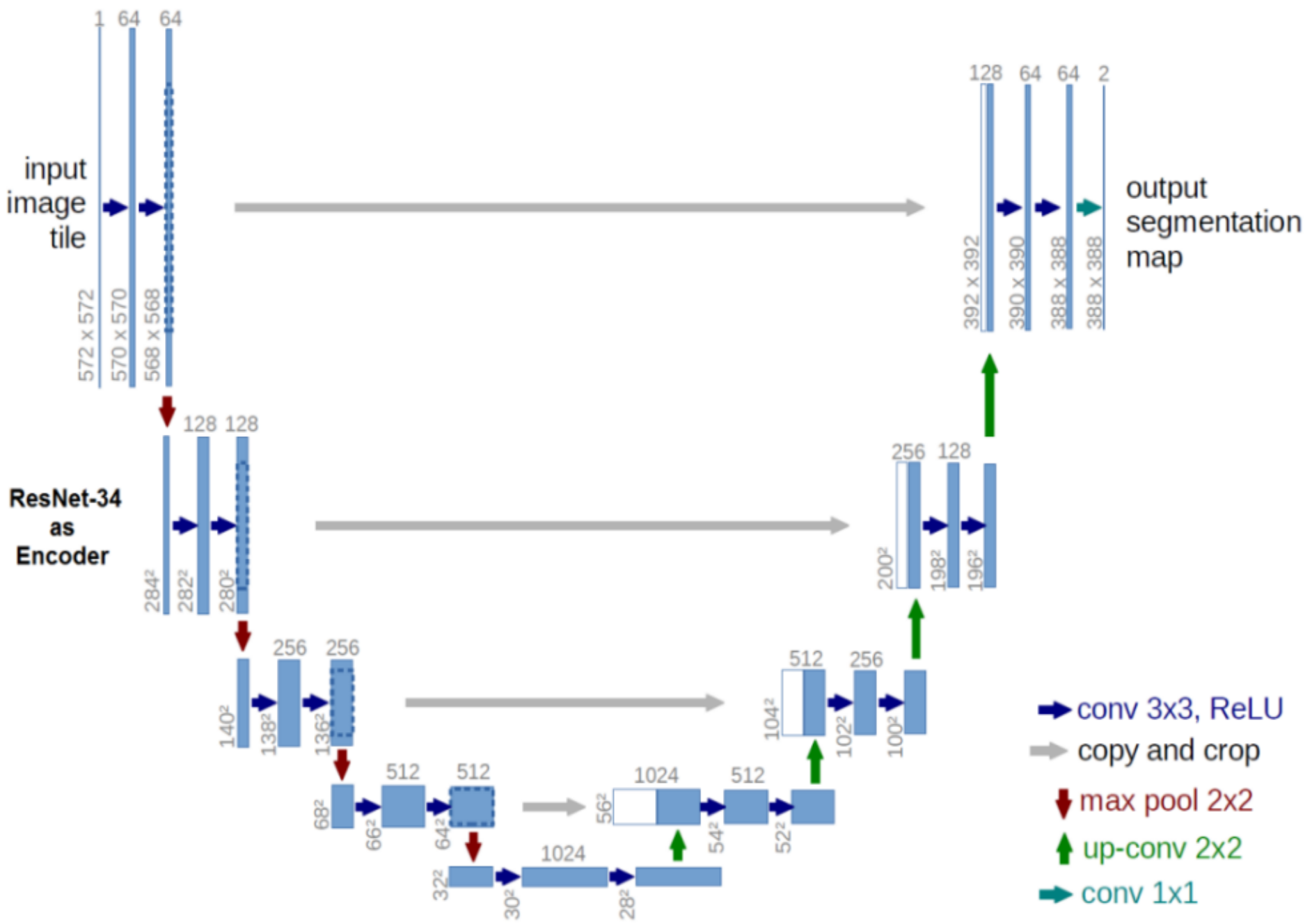

**Fig.3** Original U-net architecture for image segmentation.

### 2.3 Detecting Day and Night

In this framework, we developed a model to identify the time of day or night based on CCTV images. The model uses an average pixel value of the image to make the determination. When the average pixel value of the image surpasses 100, the model determines that it is daylight. If the pixel values fall below 100, the state is identified as nighttime. Fig.4 illustrates the application of the image mask to calculate the pixel values of the image and make the determination of the time of day or night. The image mask is used to isolate the specific region of interest in the image and the values of the pixels in that region are averaged. If the average is higher than 100, then the model considers it to be daylight. If the average is lower than 100, then the model considers it to be nighttime. The application of the image mask makes the model more accurate as it focuses on the region of interest and eliminates the noise from the background. This method is useful for CCTV images as it captures the environment accurately and makes the determination of the time of day or night more precise. Therefore, this model developed in this framework uses the average pixel value of the CCTV image to detect and categorize day and night based on the photographs. The application of the image mask ensures that the model is accurate and precise in its determination.

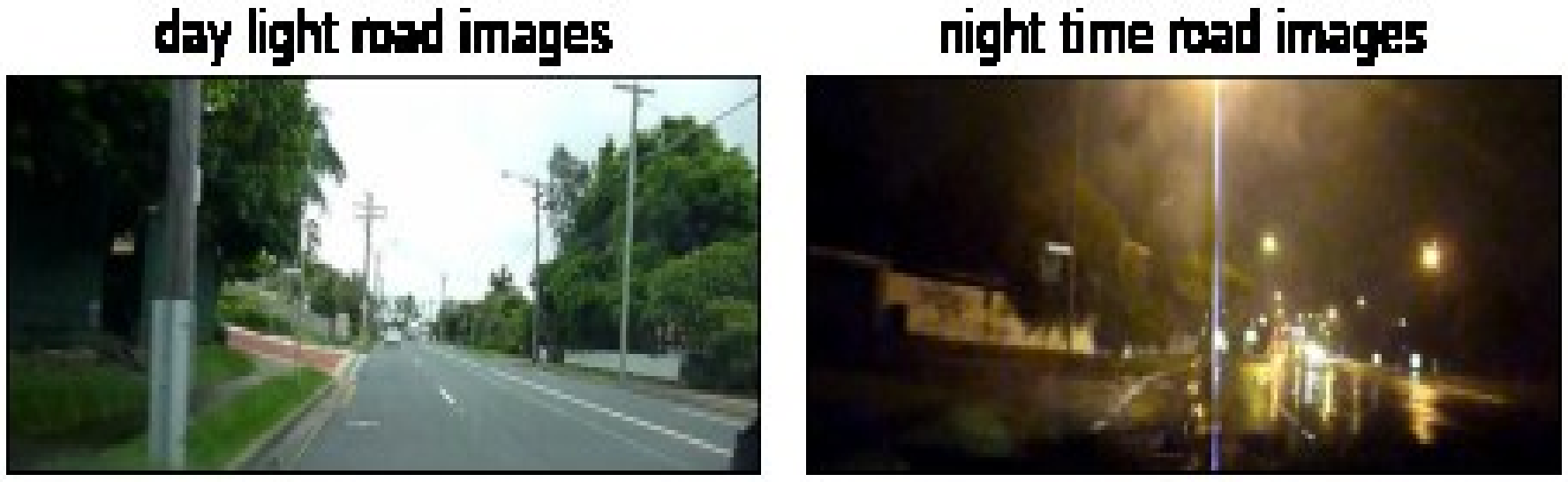

(a)

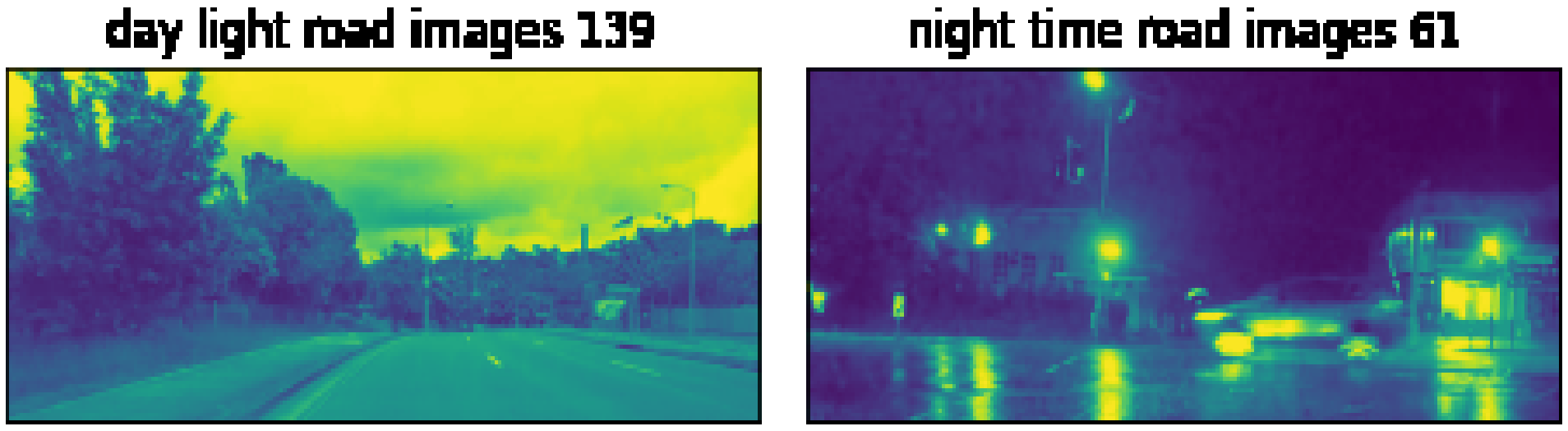

(b)

**Fig.4** Day and night classification, a. original images and b. masked images.

In today's world, advanced technologies such as geological sensors are being used in many applications. Geological sensors are used to measure physical properties such as location, temperature and other meteorological data. These sensors can be used to make educated decisions by feeding directly to the decision node.

The combination of geo-location sensors with ambient light sensors can prove to be beneficial in many situations. Geo-location sensors can help to identify the exact location of a particular object or person. On the other hand, ambient light sensors can be used to measure the brightness of the surrounding environment. This combination can be used to accurately identify and track the location of objects or people.

Consequently, depending on the regional characteristics such as climate, terrain, or flat area, suitable sensors can be chosen. For example, if the area is flat, then pressure sensors can be used to measure the pressure difference between two points. Similarly, if the area is steep, then temperature sensors can be used to measure the surrounding temperature. Thus, depending on the regional characteristics, the most suitable sensors can be chosen for effective functioning. In short, geological sensors are very useful in measuring physical properties such as location and meteorological data. They can be fed directly to the decision node to make educated judgments. Moreover, combining geo-location sensors with ambient light sensors can be beneficial in many situations. Overall, depending on regional characteristics, suitable sensors can be chosen accordingly.

### 2.4 Training Setup

As was mentioned before, the ResNet-34 model is the one that we like to use for our design. The performance of ResNet-34 as the backbone in analogous segmentation difficulties has already been demonstrated to be exceptional [28, 29]. Batch sizes can be limited to a maximum of eight, while epochs can have a maximum of twenty. The pace of learning can be slowed down to 0.0001 percent if necessary. As the objective of the model that we offer in this research is to identify these types of things, we may find that limiting the classes that are to be recognized by the model to just two, namely "vehicle" and "pedestrian," is sufficient to meet our requirements. The Softmax classifier is a useful tool for distinguishing between different kinds of objects. Because augmentation is an excellent method for boosting the quantity of data and preventing model overfitting, we may use the "albumentation" Python module to augment the training data. This would be done in order to prevent model overfitting. Due to the fact that there was just a little quantity of training data, several augmentation strategies, such as Gaussian noise, random cut, horizontal flip, etc., can be utilized to add more information to the dataset.

## 3 Results

Results of training the proposed model on the augmented dataset can be demonstrated with several evaluation matrices.

.

### 3.1 Intersection over Union (IoU) or the Jaccard Index

This is one of the most commonly used metrics in semantic segmentation. It is defined as the area of intersection between the predicted segmentation map and the ground truth, divided by the area of union between the predicted segmentation map and the ground truth:

$$\text{IoU} = J(A, B) = \frac{|A \cap B|}{|A \cup B|} \tag{1}$$

Where, A and B denote the ground truth and the predicted segmentation maps, respectively. It ranges between 0 and 1.

#### 3.1.1 Mean-IoU

A popular metric which is defined as the average Intersection over Union (IoU) over all classes is known as Mean-IoU. In other words, the average area of intersection between the predicted segmentation map and the ground truth, divided by the average area of union between the predicted segmentation map and the ground truth: It is widely used in reporting the performance of modern segmentation algorithms. From Fig.5 we can observe the mean-IoU score of our model over the whole training period was recorded as 0.65, which is pretty satisfactory according to the study [30].

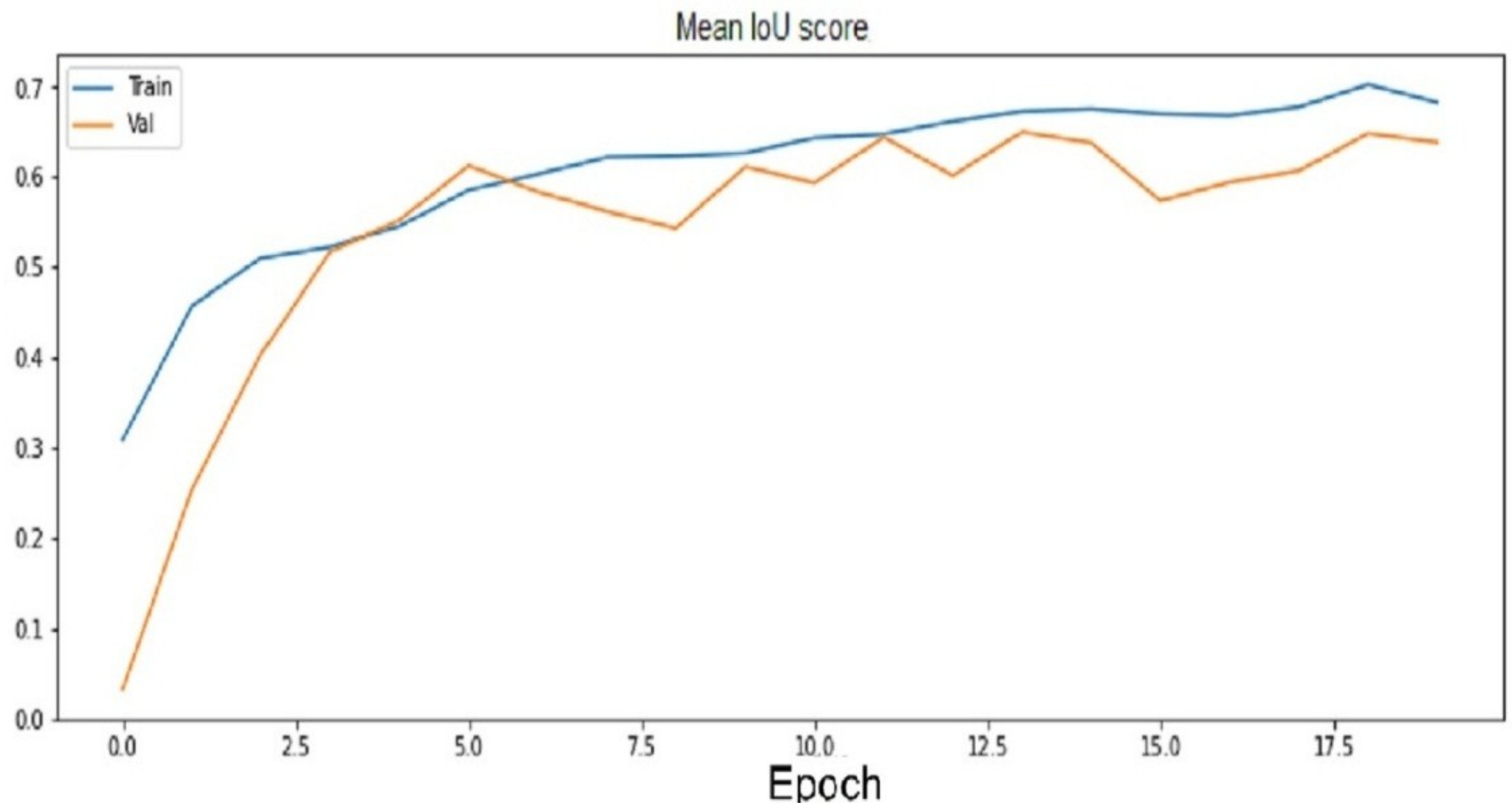

**Fig.5** Mean-IoU score of proposed model.

### 3.2 Loss

Categorical cross-entropy is used as the loss function of our model as shown in Eq. (2), which employs Softmax activations in the output layer.

$$Loss = \sum_{i=1}^{N} \sum_{j=1}^{K} t_{ij} ln\, y_{ij} \tag{2}$$

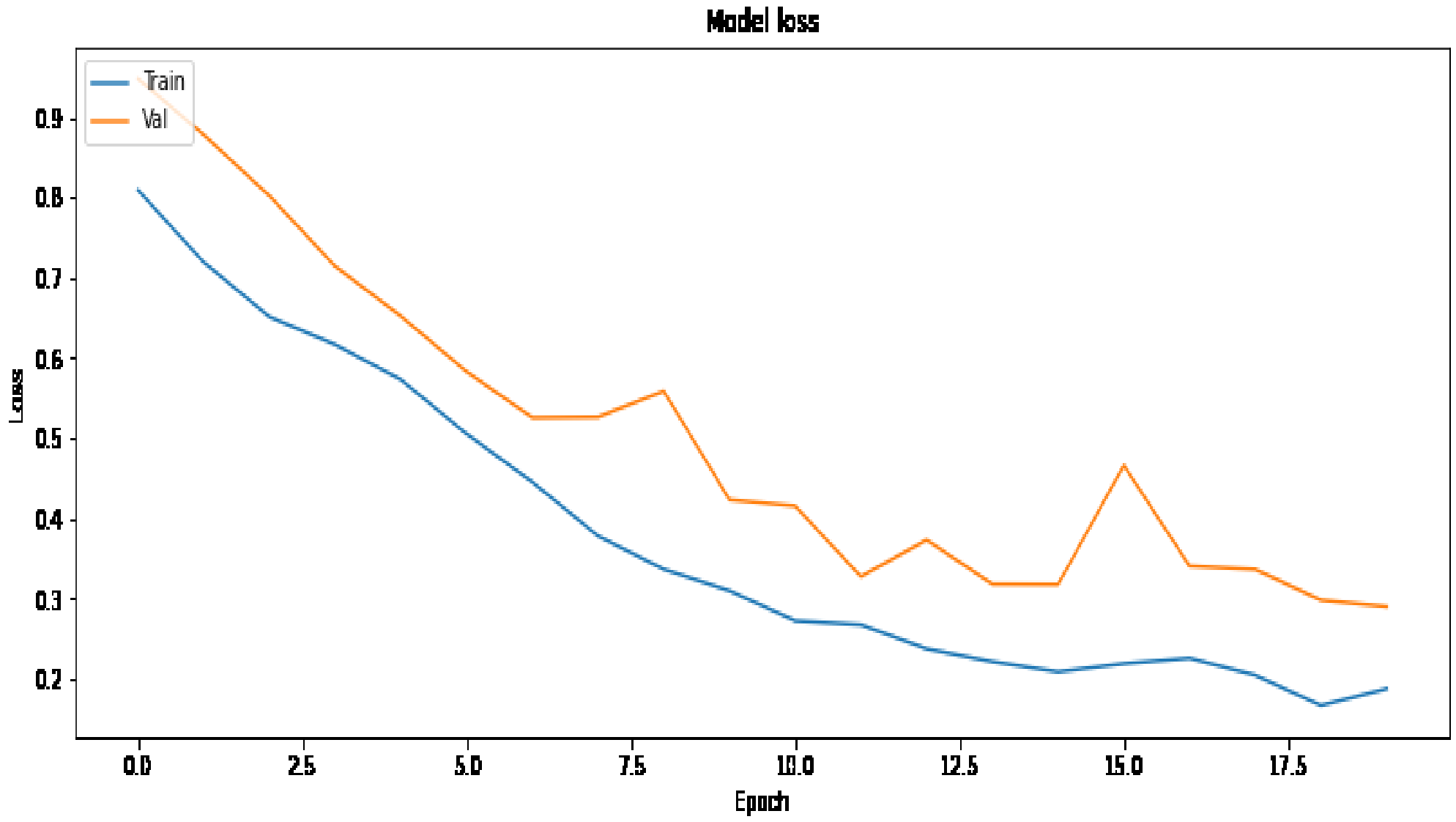

**Fig.6** Model loss over 20 epochs.

In Eq.(2), N indicates the number of image samples, K is the number of image classes, $t_{ij}$ indicates that $_i$th image sample belongs to $_j$th image class and $y_{ij}$ represents the output for image sample i for image class j.

From Fig.6 we observe the loss of our model over the twenty epochs. The loss has been decreased over the entire training period. The final loss is 0.33.

### 3.3 F1-score

Another popular metric is called the F1 score, which is defined as the harmonic mean of precision and recall:

$$\text{F1-score} = \frac{2\,(Prec\ x\ Rec\,)}{Prec + Rec.} \tag{3}$$

The mean F1-score achieved by our model is 0.72. In Fig.7, the results of the original masks and predicted masks of the proposed model are shown on three random test photos from the CamVid dataset.

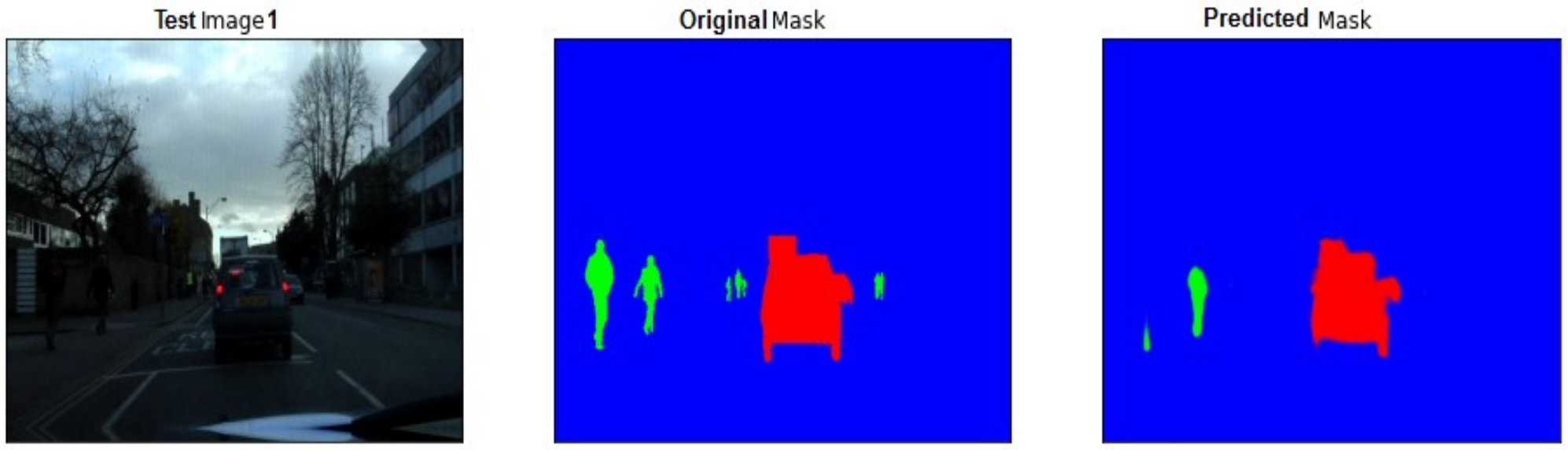

(a)

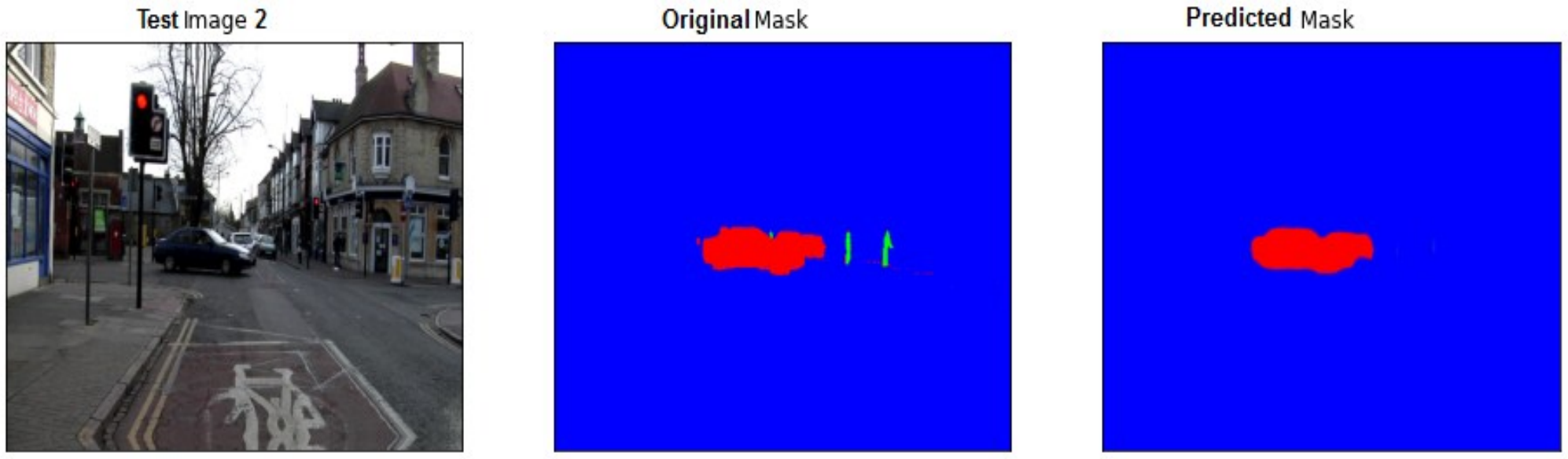

(b)

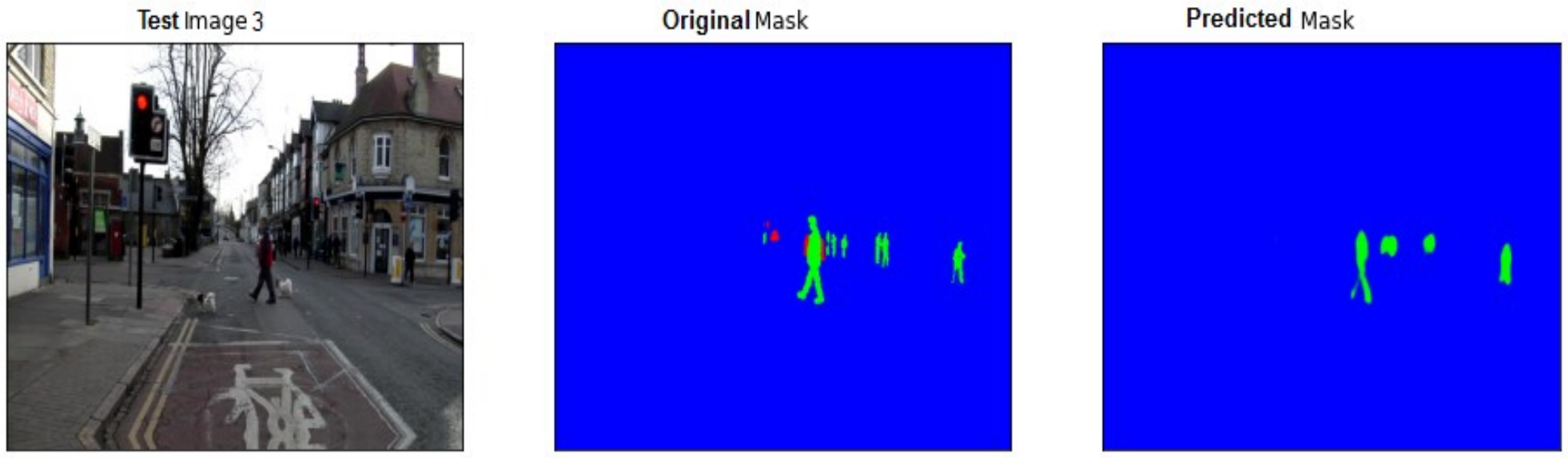

(c)

**Fig.7** Three test images and their corresponding original mask and predicted mask images are shown in (a), (b) and (c)

## 4 Discussions and Challenges

Object detection is a crucial part of any intelligent transport monitoring network. It is used to detect objects, such as pedestrians and vehicles, within a given scene. Without this technology, it would be difficult to implement the autonomous activation of streetlights too. This is why it is important to choose the right object detection model for the task at hand.

In this study, we chose to use U-net for our object detection system. U-net is an encoder-decoder type of architecture that can be used for segmentation as well as object detection. It is advantageous because it can be used to handle a wide range of issues associated with a complex and comprehensive framework of the intelligent transport monitoring network.

For the purpose of this study, we focused primarily on the ability of our model to detect walkers and cars and to distinguish between day and night. We were able to achieve good results, proving the applicability of our model for this task. In the future, other basic object detection models, such as YOLO, Single Shot Detector (SSD), Fast R-CNN, and others, can be chosen for use in the system. This is because each model has its own strengths and weaknesses, and there is no single model that is best for all tasks.

Despite the fact that this work presents a sophisticated streetlight-controlling mechanism with better precision even when employing affordable components, more research is still required to validate its practical applicability. This is especially important because real-world conditions may be more complex than those found in a laboratory setting.

To this end, further research should be conducted to test the system in more realistic environments and to evaluate its performance under different conditions. This could include testing the system in different lighting conditions, in the presence of different types of objects, and in different sizes of scenes. Such research will help to ensure that the system is capable of tackling more intricate circumstances in real-world conditions. This study presented in this paper enables the autonomous activation of streetlights with U-net for object detection. The system was found to be successful in detecting walkers and cars, and in distinguishing between day and night. However, further research is necessary to ensure that the system can be applied in a practical setting.

## 5 Limitations

Installing CNN-based intelligent streetlight management, along with smart CCTV cameras and semantic segmentation, imposes a number of constraints on the system's users, who are primarily a governmental body or agency. Smart CCTV cameras are a direct outcome of integrating these technologies into the system. Three of the most important factors to take into account and give careful consideration are the initial setup cost of the necessary hardware and software, the amount of processing power required to analyze the data from the surveillance camera, and the precision of the segmentation model. Both the hardware and the software that the company is purchasing are likely to cost the customer a sizable sum of money. However, it's still possible that this will not end up being the case. Software required to evaluate data from smart CCTV cameras is also expected to be expensive. This is an important factor to make.

Because of the high-end nature of the cameras' constituent parts, this will drive up their already high price tags.

Depending on the specifics of the setup, the processing of the data also requires the use of a robust computer, which may be fairly expensive. In addition, it is possible that the price of the training data required to train the semantic segmentation model would be rather high. This is because extensive amounts of training data are needed to successfully train the model as training the model requires both a large amount of data and a substantial amount of time. As the cost depends on how much of the dataset is really being used. This is due to the fact that training the model is essential before it can be put to use. However, this creates an additional problem since sufficient computing power is required to properly handle the data received from the CCTV camera.

The way in which this information is managed must adhere to certain norms established by the business world. Given that the data require some sort of prior processing before the semantic segmentation model can make use of them, it stands to reason that this condition would emerge. Given the massive number of computational resources required for this pre-processing, it is probable that completing it will take a long time. In addition, if there is a large amount of data that needs to be processed, finishing the processing of the data obtained from the CCTV camera might take a very long time. Furthermore, it takes quite some time to process the data acquired from the CCTV camera. The reason for this is that processing the data takes a long time.

To summarize, the semantic segmentation model has certain flaws, including a lack of acceptable precision, albeit this is by no means the least of the problems. The semantic segmentation model has these and other problems because of it. This is one of the drawbacks that arises. The degree to which the accuracy of the model is enhanced as a result of the training it got is influenced by the quality of the training data that was used during the process of training the model. Accuracy of the model will rapidly decrease if poor quality data are utilized during training. This is due to the fact that the data used in the model has a direct impact on its correctness. This is the natural consequence of the situation, thus this is what will happen. In this case, the link between the two concepts is exactly the same as if they were one and the same. Not only that, but the amount of data utilized during the model's training phase might also influence how well it performs in the real world. Given this reality, it is crucial to ensure that the training data are of high quality and that there is sufficient data in the dataset to complete the current task. There are several issues with the implemented CNN-based Intelligent Streetlight Management Using Smart CCTV Camera and Semantic Segmentation that need to be taken into account. We need to think about these drawbacks. The sentence that follows explains these defects and deficiencies. Some of the most essential considerations are the cost of the necessary hardware and software, the amount of processing power required to evaluate the data from the CCTV camera, and the level of accuracy offered by the segmentation model. Despite these obstacles, it is possible that the management of the lighting might benefit from implementing this method if it is carried out properly. The reason for this is because it has the potential to reduce power consumption while also contributing to a higher degree of safety. This is due to the fact that it might have a role in reducing overall power consumption.

## 6 Conclusions

The use of deep convolutional neural networks (CNNs) based on the U-net is an effective approach for dimming street lights based on the time of day and recognizing cars and pedestrians using semantic segmentation of CCTV images. This approach is much more cost-efficient, energy-saving and does not require physical sensors as compared to existing or conventional intelligent solutions. This research provides an advanced and efficient streetlight control mechanism with improved accuracy even when using affordable components. However, further research is needed to fully address this issue and develop more effective solutions. For example, researchers could explore the use of various image processing and machine learning algorithms to improve the accuracy of the detection and recognition of objects in the CCTV images. Additionally, future research can also be carried out to improve the streetlight control system to make it more adaptive to the environmental conditions and the changing traffic patterns. Furthermore, the current infrastructure for managing streetlights can be adapted for both streetlight control and street surveillance. This will help video surveillance systems to handle more complex scenarios in real-world circumstances. For instance, the system could be used to detect traffic congestions, monitor the flow of traffic, and detect and report any suspicious activities.

**Acknowledgments**

The first author conceptualized the paper. All authors contributed to the article and approved the submitted version.

Code and data available at:
https://github.com/Sakibsourav019/Streetlight_control_Multiclass_Segmentation_-Camvid-_Unet_Keras

## References

[1] R. Carvalho Barbosa, M. Shoaib Ayub, R. Lopes Rosa, D. Zegarra Rodríguez, and L. Wuttisittikulkij, "Lightweight PVIDNet: A priority vehicles detection network model based on deep learning for intelligent traffic lights," *Sensors,* vol. 20, p. 6218, 2020.

[2] A. Abdullah, S. H. Yusoff, S. A. Zaini, N. S. Midi, and S. Y. Mohamad, "Smart street light using intensity controller," in *2018 7th International Conference on Computer and Communication Engineering (ICCCE)*, 2018, pp. 1-5.

[3] M. C. V. S. Mary, G. P. Devaraj, T. A. Theepak, D. J. Pushparaj, and J. M. Esther, "Intelligent energy efficient street light controlling system based on IoT for smart city," in *2018 international conference on smart systems and inventive technology (ICSSIT)*, 2018, pp. 551-554.

[4] M. Islam, K. Y. Fariya, M. Tanim, T. Hossain, T. I. Talukder, and N. A. Chisty, "IoT-based smart street light for improved road safety," in *Smart Trends in Computing and Communications*, ed: Springer, 2022, pp. 377-390.

[5] K. Arun Bhukya, S. Ramasubbareddy, K. Govinda, and A. S. T Srinivas, "Adaptive mechanism for smart street lighting system," in *Smart Intelligent Computing and Applications*, ed: Springer, 2020, pp. 69-76.

[6] P. Kumar, "IOT based automatic street light control and fault detection," *Turkish Journal of Computer and Mathematics Education (TURCOMAT),* vol. 12, pp. 2309-2314, 2021.

[7] R. Sujatha, J. Gitanjali, R. P. Kumar, M. Ali, B. Pathy, and J. M. Chatterjee, "Automatic Street Light Control Based on Pedestrian and Automobile Detection," in *Information Security Handbook*, ed: CRC Press, 2022, pp. 201-212.
[8] S. M. Sorif, D. Saha, and P. Dutta, "Smart street light management system with automatic brightness adjustment using bolt IoT platform," in *2021 IEEE International IOT, Electronics and Mechatronics Conference (IEMTRONICS)*, 2021, pp. 1-6.
[9] P. Maheshwari, M. Agrawal, and V. Goyal, "Design and Analysis of Smart Automatic Street Light System," in *Proceedings of International Conference on Big Data, Machine Learning and their Applications*, 2021, pp. 151-159.
[10] S. Pattnaik, S. Banerjee, S. R. Laha, B. K. Pattanayak, and G. P. Sahu, "A Novel Intelligent Street Light Control System Using IoT," in *Intelligent and Cloud Computing*, ed: Springer, 2022, pp. 145-156.
[11] D. Kumar Saini, S. Meena, K. Choudhary, S. Bedia, A. Agarwal, and V. K. Jadoun, "Auto Streetlight Control with Detecting Vehicle Movement," in *Advances in Energy Technology*, ed: Springer, 2022, pp. 279-288.
[12] B. Kim, S. Han, and J. Kim, "Discriminative region suppression for weakly-supervised semantic segmentation," in *Proceedings of the AAAI Conference on Artificial Intelligence*, 2021, pp. 1754-1761.
[13] N. Jindal, "Copy move and splicing forgery detection using deep convolution neural network, and semantic segmentation," *Multimedia Tools and Applications,* vol. 80, pp. 3571-3599, 2021.
[14] J. Amin, M. A. Anjum, M. Sharif, S. Kadry, A. Nadeem, and S. F. Ahmad, "Liver Tumor Localization Based on YOLOv3 and 3D-Semantic Segmentation Using Deep Neural Networks," *Diagnostics,* vol. 12, p. 823, 2022.
[15] V. Zamani, H. Taghaddos, Y. Gholipour, and H. Pourreza, "Deep semantic segmentation for visual scene understanding of soil types," *Automation in Construction,* vol. 140, p. 104342, 2022.
[16] M. Rahnemoonfar, T. Chowdhury, A. Sarkar, D. Varshney, M. Yari, and R. R. Murphy, "Floodnet: A high resolution aerial imagery dataset for post flood scene understanding," *IEEE Access,* vol. 9, pp. 89644-89654, 2021.
[17] C. Sakaridis, D. Dai, and L. Van Gool, "ACDC: The adverse conditions dataset with correspondences for semantic driving scene understanding," in *Proceedings of the IEEE/CVF International Conference on Computer Vision*, 2021, pp. 10765-10775.
[18] Q. M. Rahman, N. Sünderhauf, P. Corke, and F. Dayoub, "Fsnet: A failure detection framework for semantic segmentation," *IEEE Robotics and Automation Letters,* vol. 7, pp. 3030-3037, 2022.
[19] S. Frizzi, M. Bouchouicha, J. M. Ginoux, E. Moreau, and M. Sayadi, "Convolutional neural network for smoke and fire semantic segmentation," *IET Image Processing,* vol. 15, pp. 634-647, 2021.
[20] Z. Wang, C. Zheng, J. Yin, Y. Tian, and W. Cui, "A Semantic Segmentation Method for Early Forest Fire Smoke Based on Concentration Weighting," *Electronics,* vol. 10, p. 2675, 2021.
[21] Z. Wang, P. Yang, H. Liang, C. Zheng, J. Yin, Y. Tian*, et al.*, "Semantic segmentation and analysis on sensitive parameters of forest fire smoke using smoke-unet and landsat-8 imagery," *Remote Sensing,* vol. 14, p. 45, 2021.
[22] D. Rashkovetsky, F. Mauracher, M. Langer, and M. Schmitt, "Wildfire detection from multisensor satellite imagery using deep semantic segmentation," *IEEE Journal of Selected Topics in Applied Earth Observations and Remote Sensing,* vol. 14, pp. 7001-7016, 2021.
[23] G. T. S. Ho, Y. P. Tsang, C. H. Wu, W. H. Wong, and K. L. Choy, "A Computer Vision-Based Roadside Occupation Surveillance System for Intelligent Transport in Smart Cities," *Sensors,* vol. 19, p. 1796, 2019.
[24] D. Feng, C. Haase-Schütz, L. Rosenbaum, H. Hertlein, C. Glaeser, F. Timm*, et al.*, "Deep multi-modal object detection and semantic segmentation for autonomous driving: Datasets, methods, and challenges," *IEEE Transactions on Intelligent Transportation Systems,* vol. 22, pp. 1341-1360, 2020.

[25] B. Chen, C. Gong, and J. Yang, "Importance-aware semantic segmentation for autonomous vehicles," *IEEE Transactions on Intelligent Transportation Systems,* vol. 20, pp. 137-148, 2018.

[26] M. Colley, B. Eder, J. O. Rixen, and E. Rukzio, "Effects of semantic segmentation visualization on trust, situation awareness, and cognitive load in highly automated vehicles," in *Proceedings of the 2021 CHI Conference on Human Factors in Computing Systems*, 2021, pp. 1-11.

[27] O. Ronneberger, P. Fischer, and T. Brox, "U-net: Convolutional networks for biomedical image segmentation," in *International Conference on Medical image computing and computer-assisted intervention*, 2015, pp. 234-241.

[28] T. Zheng, H. Fang, Y. Zhang, W. Tang, Z. Yang, H. Liu*, et al.*, "Resa: Recurrent feature-shift aggregator for lane detection," in *Proceedings of the AAAI Conference on Artificial Intelligence*, 2021, pp. 3547-3554.

[29] A. Abedalla, M. Abdullah, M. Al-Ayyoub, and E. Benkhelifa, "The 2ST-UNet for pneumothorax segmentation in chest X-Rays using ResNet34 as a backbone for U-Net," *arXiv preprint arXiv:2009.02805,* 2020.

[30] Z. Huang, L. Huang, Y. Gong, C. Huang, and X. Wang, “Mask scoring r-cnn,” in Proceedings of the IEEE/CVF conference on computer vision and pattern recognition, 2019, Conference Proceedings, pp. 6409–6418.

[31] Li, J., Mei, X., Prokhorov, D., & Tao, D. (2016). Deep neural network for structural prediction and lane detection in traffic scene. *IEEE transactions on neural networks and learning systems*, *28*(3), 690-703.

[32] Dong, Y., Patil, S., van Arem, B., & Farah, H. (2023). A hybrid spatial–temporal deep learning architecture for lane detection. *Computer‐Aided Civil and Infrastructure Engineering*, *38*(1), 67-86.

[33] Tourani, A., Shahbahrami, A., Soroori, S., Khazaee, S., & Suen, C. Y. (2020). A robust deep learning approach for automatic iranian vehicle license plate detection and recognition for surveillance systems. *IEEE Access*, *8*, 201317-201330.

[34] Chen, R. C. (2019). Automatic License Plate Recognition via sliding-window darknet-YOLO deep learning. *Image and Vision Computing*, *87*, 47-56.

[35] Chen, C., Liu, B., Wan, S., Qiao, P., & Pei, Q. (2020). An edge traffic flow detection scheme based on deep learning in an intelligent transportation system. *IEEE Transactions on Intelligent Transportation Systems*, *22*(3), 1840-1852.

[36] Farag, M., Din, M., & Elshenbary, H. (2020). Deep learning versus traditional methods for parking lots occupancy classification. *Indonesian Journal of Electrical Engineering and Computer Science*, *19*(2), 964-973.